\documentclass[letterpaper, 10 pt, conference]{ieeeconf}  

\IEEEoverridecommandlockouts                              

\usepackage{amsmath}
\usepackage{amsfonts}
\usepackage{algorithm}
\usepackage{graphicx}
\usepackage{caption}
\usepackage{subcaption}
\usepackage{wrapfig}
\usepackage{algpseudocode}
\usepackage{hyperref}

\DeclareMathOperator*{\argmax}{arg\,max}

\newcommand{\savefootnote}[2]{\footnote{\label{#1}#2}}
\newcommand{\repeatfootnote}[1]{\textsuperscript{\ref{#1}}}

\title{\LARGE \bf
Human-Centric Active Perception for Autonomous Observation
}

\author{David Kent$^{*}$ and Sonia Chernova$^{*}$
\thanks{$^{*}$Georgia Institute of Technology, Atlanta, GA, USA
        {\tt\small \{dekent, chernova\}@gatech.edu}}%
}

\begin{document}

\maketitle
\thispagestyle{empty}
\pagestyle{empty}

\begin{abstract}


As robot autonomy improves, robots are increasingly being considered in the role of autonomous observation systems — free-flying cameras capable of actively tracking human activity within some predefined area of interest. In this work, we formulate the autonomous observation problem through multi-objective optimization, presenting a novel Semi-MDP formulation of the autonomous human observation problem that maximizes observation rewards while accounting for both human- and robot-centric costs.  We demonstrate that the problem can be solved with both scalarization-based Multi-Objective MDP methods and Constrained MDP methods, and discuss the relative benefits of each approach. We validate our work on activity tracking using a NASA Astrobee robot operating within a simulated International Space Station environment. 

\end{abstract}

\section{INTRODUCTION}

Human operations in extreme and remote environments, such as space and deep water domains, have the potential to benefit from robots with autonomous observation capabilities.  Due to their high-cost and high-risk nature, human activities in such domains are often video recorded for documentation and later analysis.  NASA, for example, collects video documentation of each experiment conducted on the International Space Station (ISS), while remote operation of underwater vehicles is similarly recorded.  As robot autonomy improves, robots are increasingly being considered in the role of \textit{autonomous observation systems} — free-flying cameras capable of actively tracking human activity within an area of interest.  Example systems include the NASA Astrobee \cite{smith2016astrobee} and European Space Agency CIMON \cite{CIMON} robots developed for the ISS, as well as autonomous camera robots being considered for underwater exploration \cite{ishida2012marker}.
 
While existing robot hardware offers capable candidates for autonomous observation systems, the autonomous observation problem itself is complex and largely unsolved.  Autonomous observation of humans moving in 3D space is challenging due to the proliferation of viewpoints required to cover unconstrained humans in 6-DOF environments.  Adding further challenge, the robot should act as a passive observer, causing minimal distraction to the human subject from both collisions and visual and auditory disturbance.

\begin{figure}[tb]
\centering
\includegraphics[width=0.485\textwidth]{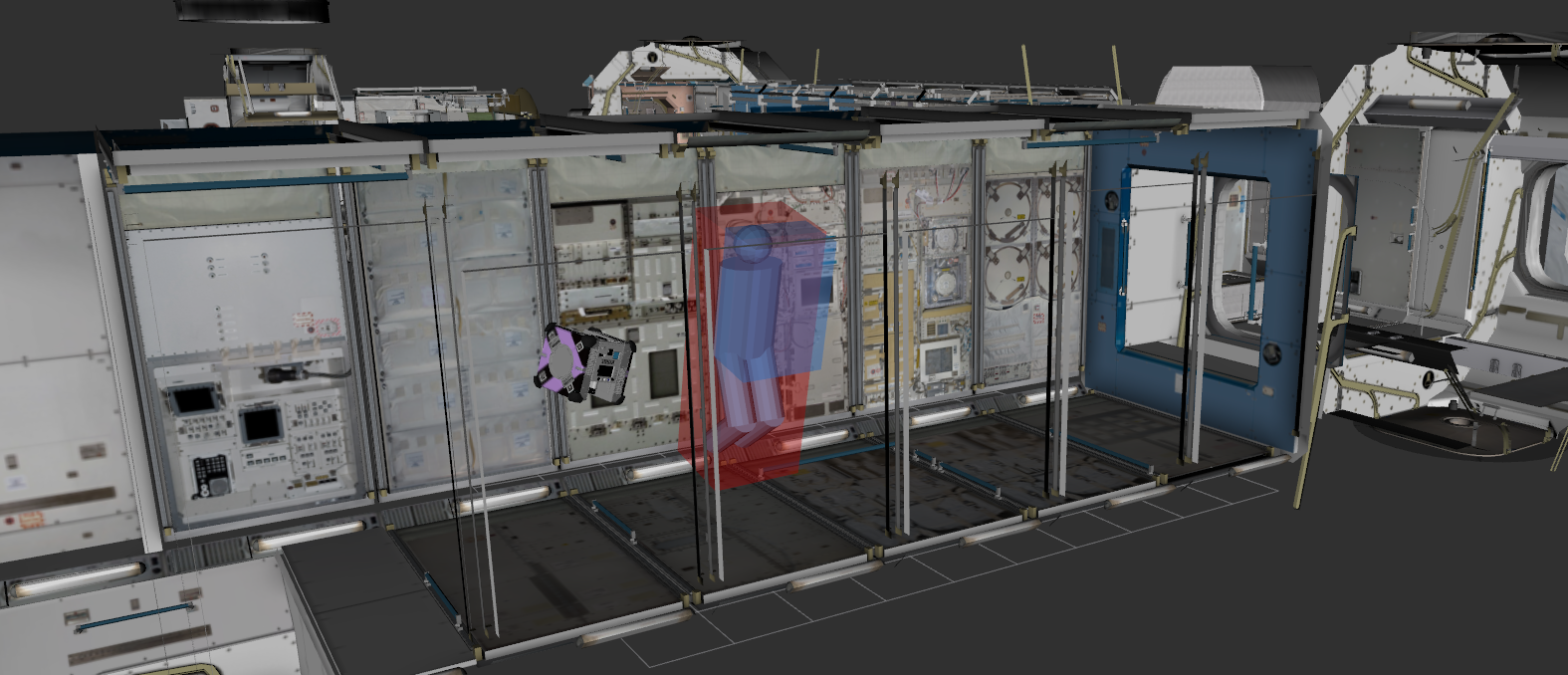}
\caption{The Astrobee platform in a module of the ISS, simulated in Gazebo and visualized in rviz.}
\label{fig:sim}
\vspace{-0.2cm}
\end{figure}

In this work, we formulate the autonomous observation problem through multi-objective optimization, as the problem requires balancing observation rewards with human-centric costs and the limitations of the robot.  We base our problem formulation on Markov Decision Processes (MDPs), as they are effective at representing sequential decision problems while accounting for the robot's transition dynamics.  We validate our approach on activity tracking using a simulated model of the Astrobee robot operating within a simulated ISS environment developed by NASA (Figure \ref{fig:sim}).

Our work makes the following contributions.  First, we show that the autonomous human observation problem can be formalized as a Semi-MDP that maximizes observation rewards while accounting for both human- and robot-centric costs.  Second, we demonstrate how the problem can be solved with both scalarization-based Multi-Objective MDP (MOMDP) methods and Constrained MDP (CMDP) methods.  Last, we discuss the two methods' relative benefits and drawbacks, supported by experimental results from performing Astrobee viewpoint planning for a set of tasks on the ISS under different sets of task constraints.  Our results show that while both of the techniques we present succeed in optimally solving the task, the underlying characteristics of these methods highlight important tradeoffs in their potential application.  The CMDP's constraint-based formulation allows for far more transparent parameter setting, eliminating the need for pre-trial simulation or run-throughs of the experimental scenario.  The MOMDP approach, however, is far more computationally efficient, and better suited for domains in which the observation task and the environment remain relatively unchanged and efficient computation is needed. 

\section{RELATED WORK}

In this section, we situate our work within the field of active perception, and discuss relevant multi-objective optimization approaches to autonomous human observation. 

\subsection{Active Perception}
Bajscy et al. define active perception according to the active pentuple \textit{why}, \textit{what}, \textit{when}, \textit{where}, \textit{how} \cite{bajcsy2018revisiting}.  Situating our problem within this definition, our active agent is particularly concerned with \textit{when} and \textit{where} to position itself to maximize the expected visual coverage of a target agent which follows its own trajectory.  Further, as a mobile robot, the positioning decisions our agent can make are constrained by \textit{how} the agent can move.  In the context of Chen et al.'s categorization of active vision tasks, our problem is most similar to the surveillance problem and the object search and tracking problem \cite{chen2011active}.

There is a large body of work on using mobile robots for surveillance, although the approaches typically focus on achieving high area coverage rather than on human tracking \cite{chen2011active}.  Prior work solves for optimal mobile camera positioning for individual optimal viewpoints with respect to field-of-view and resolution constraints, for both coverage of key regions \cite{nilsson2009towards} and target tracking \cite{bodor2005mobile}.  Schroeter et al. add lighting constraints as an extension \cite{schroeter2009autonomous}.  Other work has formulated mobile robot surveillance according to environmental and human factors by combining them into a threat profile, biasing mobile robot surveillance coverage towards areas of higher threat \cite{ma2009matching}.  The threat measure is treated as constant, however, as it changes on a large time scale, whereas the effects of our human-centric constraints change rapidly as the subject moves through the environment.  Both the self-organizing map algorithm \cite{best2016multi} and the randomized algorithm for informative path planning \cite{arora2017randomized} optimize location observation rewards under travel budgets, but they assume temporally-static targets and environments and are therefore not suited to observing moving subjects.

Much of the object search and tracking literature explores the inherent tradeoff between search and tracking \cite{wang2008awareness, sung2017algorithm}.  We assume the search problem is solved, and are concerned instead with the quality of observations collected during tracking, although the search and tracking tradeoff is relevant to situations where the human subject's location is not known a priori.  The online informative path planning algorithm maximizes classification probabilities under a travel budget \cite{popovic2017online}, but as with surveillance, it is designed more for coverage than tracking.  The aerial social force model \cite{garrell2017aerial} combines attractive and repulsive forces to keep a UAV near a human subject while avoiding obstacles and pedestrians, but as a reactive approach it can be myopic.

\subsection{Multi-Objective Optimization}
The fields of planning and multi-objective optimization provide relevant approaches to the autonomous human observation problem.  Many active perception tasks deal with planning under partial observability, which Partially Observable MDPs (POMDPs) directly address.  POMDPs have been successfully implemented for path planning on real robot systems \cite{ragi2013uav}, and even for planning with collaborators \cite{chen2016pomdp}, although they can be challenging to implement tractably for real-world tasks.  In this work, we focus on fully observable environments, but consider POMDPs for future extensions.

Multi-objective optimization provides a spectrum of approaches for handling rewards and costs within MDP frameworks.  The first approach we consider is scalarization, in which the set of rewards and costs are combined into a single objective by a (typically linear) scalarization function \cite{roijers2013survey}.  Scalarization also extends to combining task- and belief-based rewards and costs in POMDPs \cite{eck2012evaluating}.  The biggest drawback is the difficulty in correctly tuning the weights of a linear scalarization function, or in selecting a different scalarization function that fully captures the relationship between all of a problem's objectives.  Some approaches solve for all combinations of the weights \cite{lizotte2012linear}, but still require a mechanism for weight selection at runtime.

An alternative approach to scalarization is to optimize a primary objective, in our case the observation reward, while treating the costs as constraints.  The GUBS formulation optimizes a single objective with a cost tradeoff for Goal-Directed MDPs \cite{freire2017gubs}.  Goal-directed behavior is not suitable for our approach, as we aim to optimize observation over the full trajectory of the human subject.  Similarly, Constrained MDP (CMDP) methods have been shown to meet a single mission objective while satisfying cost constraints formulated as linear temporal logic (LTL) subgoals \cite{ding2014hierarchical, feyzabadi2016multi}.  Our constraints do not fit well into LTL, but the costs-as-constraints approach afforded by CMDPs naturally incorporates human- and robot-centric costs into our problem formulation.


\section{PROBLEM FORMULATION}
In this section, we formalize the general autonomous human observation problem as an SMDP with an associated set of cost functions, followed by an instantiation of the problem for our Astrobee case study.

\subsection{General Problem Formulation}
We define the autonomous observation problem as a Semi-Markov Decision Process (SMDP) \cite{hu2007markov} with the components
\begin{equation}
\{S, A(s), p(s'|s,a), p(\tau|s,a,s'), r(s,a,s',\tau)\},
\end{equation}
defined as follows:
\begin{itemize}
\item $s_t \in S$ is a state in the state space consisting of the robot state and the human subject's pose $[x_r, x_h]$.  We assume the robot is at one of a set of waypoints $x_r.pose \in [w_0, w_1, \ldots, w_n]$.  The set can either be user-defined or calculated with viewpoint planners such as \cite{schroeter2009autonomous}.
\item $A(s)$ is the set of actions available to the robot at the current state, which must include the subset $\{hold\_pos()\} \cup \{move(w_i) | w_i \in [w_0, w_1, \ldots, w_n]\}$.
\item $p(s'|s,a)$, the state transition function, is the probability that executing action $a$ in state $s$ will result in state $s'$.
\item $p(\tau|s,a,s')$, the time duration distribution function, is the probability that transitioning from state $s$ to state $s'$ with action $a$ will take duration $\tau$.
\item $r(s,a,s',\tau)$ is the reward function.  We model this as an observation reward rate received over the period $\tau$, i.e. $r(s,a,s',\tau)=r(s,a,s')\tau$.
\end{itemize}

We define the observation reward based on subject coverage and resolution (a function of distance), as they primarily affect the image quality of the observation images \cite{bodor2005mobile}.  We calculate the reward as the expected percentage of a region-of-interest ($ROI$) covered by the robot's field of view $V_r$, scaled by the distance from the robot to the ROI center.
\begin{equation}
r(s,a,s') = \frac{1}{||ROI_{c} - x_r.pose||}\frac{||V_r \bigcap ROI||}{||ROI||}
\label{eq:reward}
\end{equation}
The ROI can be defined as a human-centric task workspace, the subject's full bounding box, or whatever area the robot's camera should capture.

Additionally, we introduce a set of constraints $c_i(s,a,s',\tau)$ to model human- and robot-centric costs.  Similar to the reward function, we accumulate costs over a time duration, i.e. $c_i(s,a,s',\tau)=c_i(s,a,s')\tau$.  The costs are as follows:
\begin{itemize}
  \item $c_0(s,a,s')$ represents potential collision between the robot and the human, which is calculated based on the distance from the robot to a bounding box around the human's workspace\savefootnote{foot:cost}{Taken together, the human-centric costs $c_0$ and $c_1$ account for human proxemics, with a hard-constrained personal space (the human workspace bounding box) surrounded by exponentially increasing distance zones \cite{edward1966hall}.}, shown in red in Figure \ref{fig:sim}.  The platform-specific parameter $\alpha_0$ controls how close to the workspace edge the robot can be.
  \begin{equation}
  c_0(s,a,s') = e^{-\alpha_0dst(x_r.pose, wkspc(x_h))}
  \label{eq:collision}
  \end{equation}
  \item $c_1(s,a,s')$ represents the degree of intrusion caused by the robot to the human, calculated based on the distance from the robot to the human's head\repeatfootnote{foot:cost}.  Note that this is in direct conflict with the observation reward.  The platform-specific parameter $\alpha_1$ controls the rate at which distance decreases the robot's intrusiveness.
  \begin{equation}
  c_1(s,a,s') = e^{-\alpha_1||x_r.pose - x_{h\_head}||}
  \label{eq:intrusion}
  \end{equation}
  \item $c_2(s,a,s')$ represents the platform-specific power consumption of each of the robot's actions.
\end{itemize}

The manner in which the cost functions are included in the problem definition depends on the planning method used, which we address in detail in Section \ref{sec:methods}.

As a final point, we note that the reward calculation and the human-centric cost calculations depend on knowing the human's exact pose, $x_h$, at all time steps, which is impossible in practice.  We overcome this issue by representing the human's trajectory as a probability distribution $\mathbf{x_h}(t)$ fit to known task data, e.g. timing and pose data collected from previous task executions.  As such we replace the exact reward and cost functions with expected reward and cost functions, calculated over a set of $N$ human trajectories $x^n_h(t)$ sampled from $\mathbf{x_h}(t)$:
\begin{equation}
\label{eq:expr}
\tilde{r}(s,a,s')=\frac{1}{N}\sum_n{r(s(x^n_h), a, s'(x^n_h))}
\end{equation}
\begin{equation}
\label{eq:expc}
\tilde{c_i}(s,a,s')=\frac{1}{N}\sum_n{c_i(s(x^n_h), a, s'(x^n_h))}
\end{equation}
Throughout the rest of the paper we use tildes to denote the use of expected reward and cost functions.

\subsection{Astrobee Case Study}

\begin{figure}[t]
\centering
\includegraphics[width=0.3\textwidth]{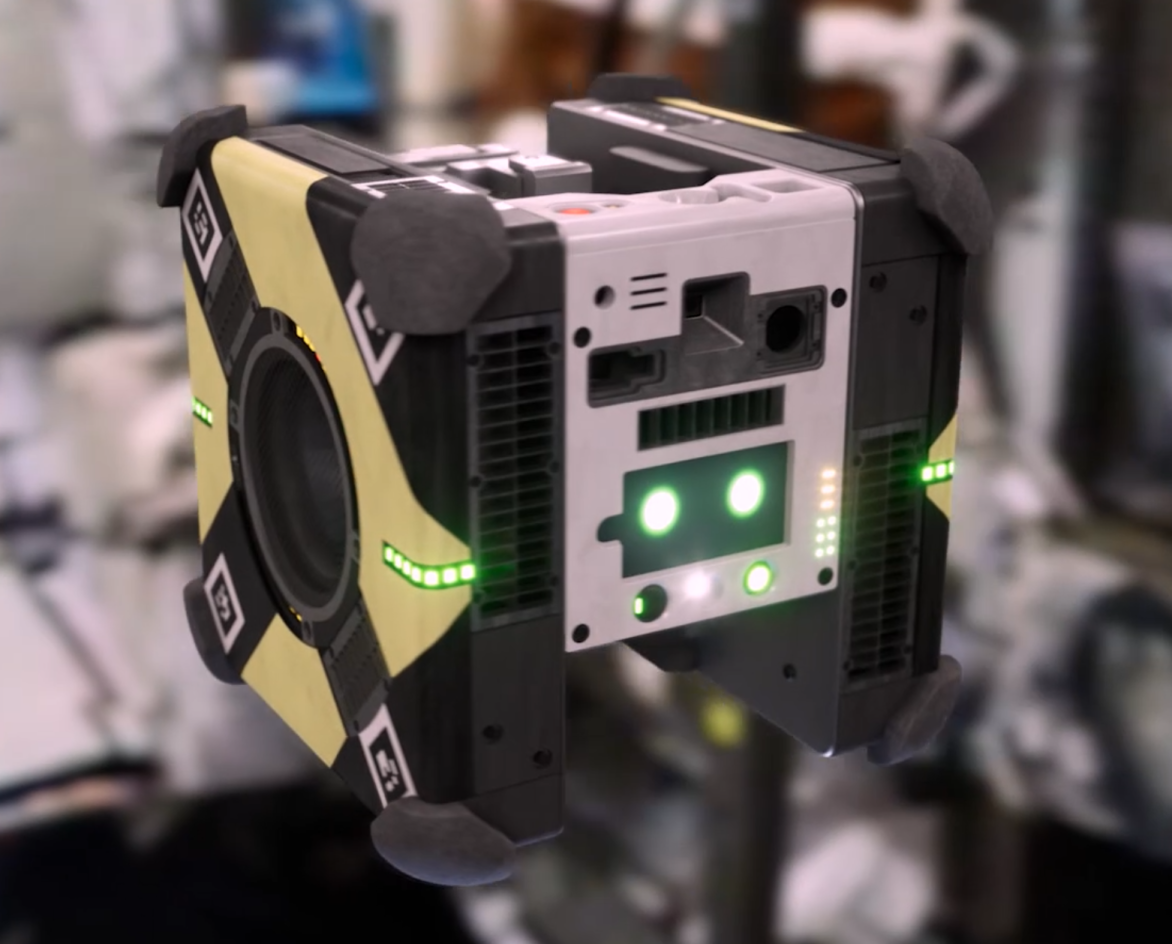}
\caption{NASA's freeflying Astrobee platform.}
\label{fig:astrobee}
\vspace{-0.3cm}
\end{figure}

In this section, we use the above formulation to define the autonomous observation problem for the NASA Astrobee robot (Figure \ref{fig:astrobee}) operating on the ISS.  Astrobee has a perching arm to attach itself to hand rails, allowing it to function as a power-saving pan-tilt camera instead of a free-flying robot.  As such, in addition to robot and human pose, our robot state $x_r$ includes a boolean $perched$ to track whether or not Astrobee is perched.  Additionally, the action set $A(s)$ includes actions for perching and unperching, available at waypoints with hand rails; the full action set becomes $\{hold\_pos(), perch(), unperch(), move(w_i)\}$.

We represent the region-of-interest used for calculating observation rewards as a rectangular prism directly in front of the person's torso and head, shown in blue in Figure \ref{fig:sim}, covering their activity workspace.  We adapt the reward function to give zero reward during $move$, $perch$, and $unperch$ actions.  This acts as a worst-case reward for Astrobee, as the robot is required to use its cameras for safe navigation.  Thus, we update the reward function of Equation \ref{eq:reward} as follows:

\begin{equation}
r(s,a,s') = \begin{cases}
\frac{1}{||ROI_{c} - x_r||}\frac{||V_r \bigcap ROI||}{||ROI||} & a\: \mathbf{is}\: hold\_pos,\\
0 & \mathbf{else}.\\
\end{cases}
\end{equation}

The cost functions require some minor extensions.  The collision cost (Equation \ref{eq:collision}) remains the same, aside from tuning $\alpha_0$.  The intrusion cost is affected by perching, in that perching reduces auditory disturbance as Astrobee can turn its fans off.  We add the indicator function $\mathtt{I}(s)$ to Equation \ref{eq:intrusion}, which returns $1$ if $perched$ is \texttt{true} and $0$ otherwise:
\begin{equation}
c_1(s,a,s') = \frac{1}{1+\mathtt{I}(s)}e^{-\alpha_1||x_r - x_{h\_head}||}.
\end{equation}
Lastly, we instantiate the power consumption cost with the lookup table:
\begin{equation}
c_2(s,a,s') = \begin{cases}
0.125 & a\: \mathbf{is}\: hold\_pos\: \mathbf{and}\: perched,\\
0.25 & a\: \mathbf{is}\: hold\_pos\: \mathbf{and}\: !perched,\\
0.5 & a\: \mathbf{is}\: perch,\\
0.5 & a\: \mathbf{is}\: unperch,\\
1.0 & a\: \mathbf{is}\: move.
\end{cases}
\end{equation}

\section{PLANNING METHODS}
\label{sec:methods}
We present two approaches to solve the SMDP with associated cost functions:  Multi-Objective MDPs (MOMDPS) with scalarization functions, and Constrained MDPs (CMDPs).  In both cases, we solve the SMDP over a finite horizon with undiscounted rewards, to optimize for total accumulated observation rewards over the fixed duration of the observation subject's task.

\subsection{MOMDPs with Scalarization}
Our first approach is to solve the SMDP by reducing our model to an MOMDP and performing backwards induction~\cite{hu2007markov} over a scalarized objective function~\cite{roijers2013survey} that combines the reward and costs.  First, we define reward and cost functions calculated for only a state and action by taking expectations over the resulting states and action durations:
\begin{equation}
\label{eq:r}
\tilde{r}(s,a) = \sum_{s'}{\left[p(s'|s,a)\sum_{\tau}{p(\tau|s,a,s')\tilde{r}(s,a,s')\tau}\right]}
\end{equation}
\begin{equation}
\label{eq:c}
\tilde{c}_i(s,a) = \sum_{s'}{\left[p(s'|s,a)\sum_{\tau}{p(\tau|s,a,s')\tilde{c}_i(s,a,s')\tau}\right]}.
\end{equation}

We then combine multiple objective functions $\mathbf{V}(s,a)$ (a vector containing all of the objectives, in our case $[\tilde{r}(s,a), \tilde{c}_0(s,a), \tilde{c}_1(s,a), c_2(s,a)]$) into a single objective function according to a set of weights $\mathbf{w}$ using a scalarization function $f(\mathbf{V}(s,a), \mathbf{w})$.  We use a linear scalarization function that treats rewards as positive and costs as negative:
\begin{equation}
\begin{split}
f(\mathbf{V}(s,a), \mathbf{w})= & \mathbf{V}(s,a) \cdot \mathbf{w},\\
& \mathbf{w}=[w_{\tilde{r}}, -w_{\tilde{c}0}, -w_{\tilde{c}1}, -w_{c2}].
\end{split}
\end{equation}
For our application, the weights are selected by an expert in advance (although weight selection may not be straightforward, see Section \ref{sec:results}).

We then perform backwards induction, marginalizing out $\tau$ as in Equations \ref{eq:r} and \ref{eq:c}, to determine optimal action selection and the utilities of each state under the optimal policy for each time step, using $f(s,a) = f(\mathbf{V}(s,a), \mathbf{w})$ as the reward:
\begin{equation}
\begin{split}
u^*(s(t))&={}\max_{a\in A(s)}f(s,a) +{} \\
&\sum_{s'\in S}{p(s'|s,a)\sum_{\tau}{p(\tau|s,a,s')u^*(s'(t+\tau))}}
\end{split}
\label{eq:bi1}
\end{equation}
\begin{equation}
\begin{split}
a^*(s(t))&={}\argmax_{a\in A(s)}f(s,a) +{} \\
&\sum_{s'\in S}{p(s'|s,a)\sum_{\tau}{p(\tau|s,a,s')u^*(s'(t+\tau))}}.
\end{split}
\label{eq:bi2}
\end{equation}

\subsection{CMDPs}
Alternatively, the problem can be framed as a Constrained MDP\footnote{We note that CMDPs are a subset of MOMDPs, but for brevity we refer to the two methods we consider as MOMDPs (with linear scalarization) and CMDPs.} \cite{altman1999constrained}, represented by the following tuple:
\begin{equation}
\{S, s_0, A(s), p(s'|s,a), \tilde{r}(s,a), \tilde{\mathbf{c}}(s,a), \mathbf{d}\},
\end{equation}
where $\tilde{c}_i(s,a) \in \tilde{\mathbf{c}}$ is a cost function (i.e. Equation \ref{eq:c}), and $d_i \in \mathbf{d}$ is a constraint value associated with $\tilde{c}_i$.  The goal of a CMDP is to maximize the expected total reward subject to a set of constraints defined by the expected total costs:
\begin{equation}
\begin{split}
\max_{\pi}{} & u^\pi_{\tilde{r}}(s_0)= \mathbb{E}_{\pi}\left[\sum_{t=0}^{N}\tilde{r}(s_t,a_t)|s_0\right] \\
s.t.\, & u_{\tilde{c}}^\pi(s_0) = \mathbb{E}_\pi\left[\sum_{t=0}^N{\tilde{c}_k(s_t,a_t)|s_0}\right] \leq d_k \:\:\:\: \forall k.
\end{split}
\end{equation}
As such, instead of specifying a set of weights for a scalarization function, the CMDP requires the user to specify a set of hard thresholds bounding the expected accumulated costs.  

The CMDP is solved by reformulating the above problem as a linear program where we are solving for the variables $y_{s,a}$, which represent weighted occupancy for each state-action pair (see \cite{altman1999constrained} for a full derivation of this method):
\begin{equation}
\begin{split}
\max_{y_{s,a}}{} & {\sum_{s,a}{\tilde{r}(s,a)y_{s,a}}}\\
s.t.  & \sum_{a'}{y_{s',a'}}=\delta(s_0,s')+\sum_{s,a}{T(s'|s,a)y_{s,a}} \:\:\:\: \forall s'\\
& \sum_{s,a}{\tilde{c}_k(s,a)y_{s,a}} \leq d_k \:\:\:\: \forall k\\
& y_{s,a} \geq 0 \:\:\:\: \forall s,a.
\end{split}
\end{equation}
When an optimal $y^*_{s,a}$ is found, we recover a policy as follows:
\begin{equation}
\pi^*(a|s)=p(a|s)=\frac{y^*_{s,a}}{\sum_{a'}{y^*_{s,a'}}},
\label{eq:lp}
\end{equation}
with the expected utilities given as:
\begin{equation}
u^*_{\tilde{r}}(s_0,\mathbf{d}) = \sum_{s,a}{\tilde{r}(s,a)y^*_{s,a}}.
\end{equation}

\section{EVALUATION}
\label{sec:results}

\begin{figure}
\centering
\begin{subfigure}{.44\textwidth}
	\centering
	\includegraphics[width=\textwidth]{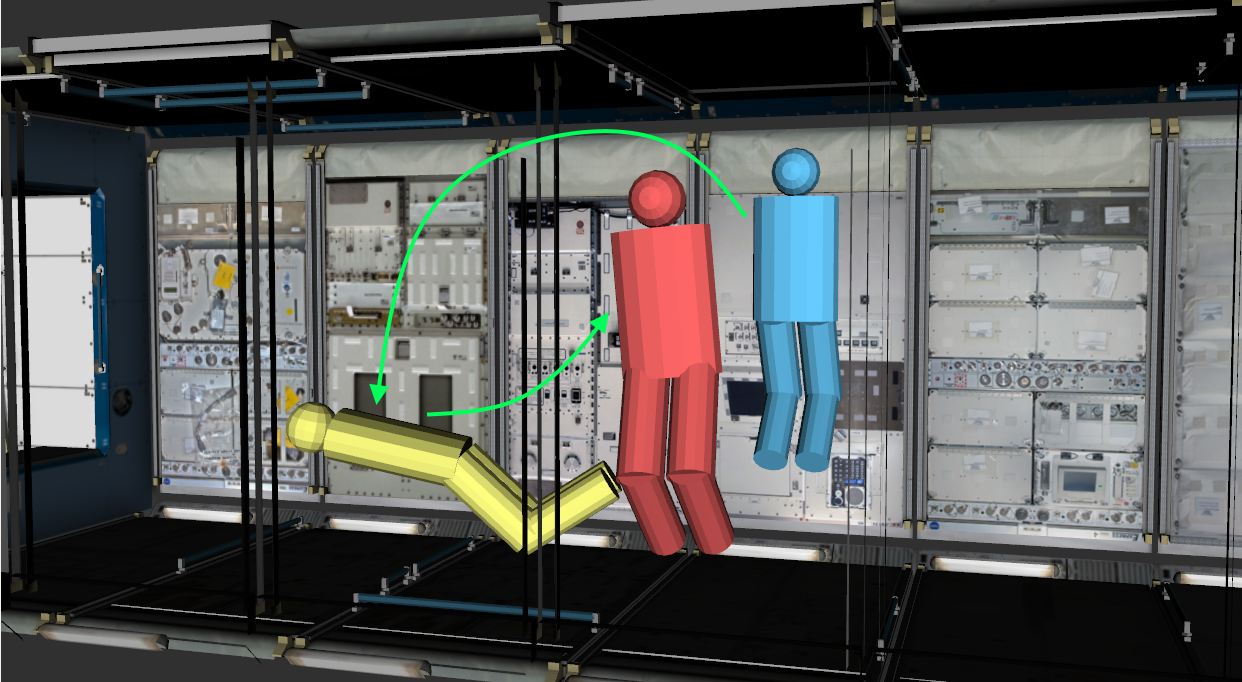}
	\vspace{-.5cm}
	\caption{Experiment task.}
	\label{fig:experiment}
\end{subfigure}
\par\medskip
\begin{subfigure}{.44\textwidth}
	\centering
	\includegraphics[width=\textwidth]{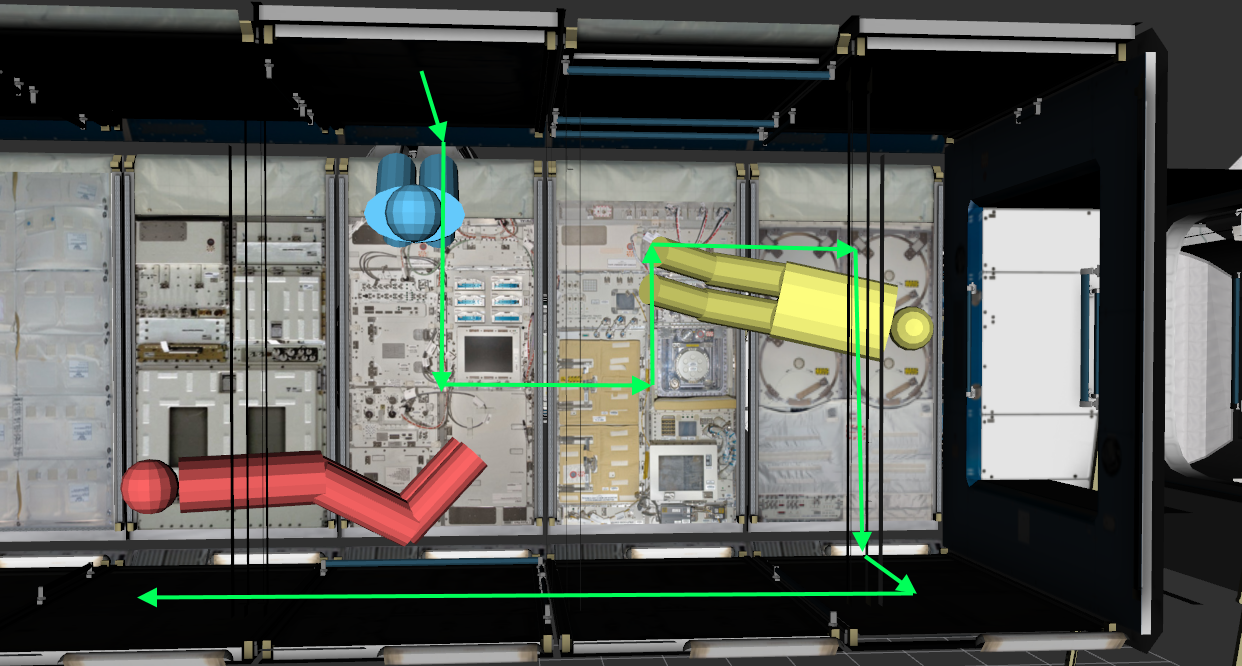}
	\vspace{-.5cm}
	\caption{Inspection task.}
	\label{fig:inspection}
\end{subfigure}
\par\medskip
\begin{subfigure}{.44\textwidth}
	\centering
	\includegraphics[width=\textwidth]{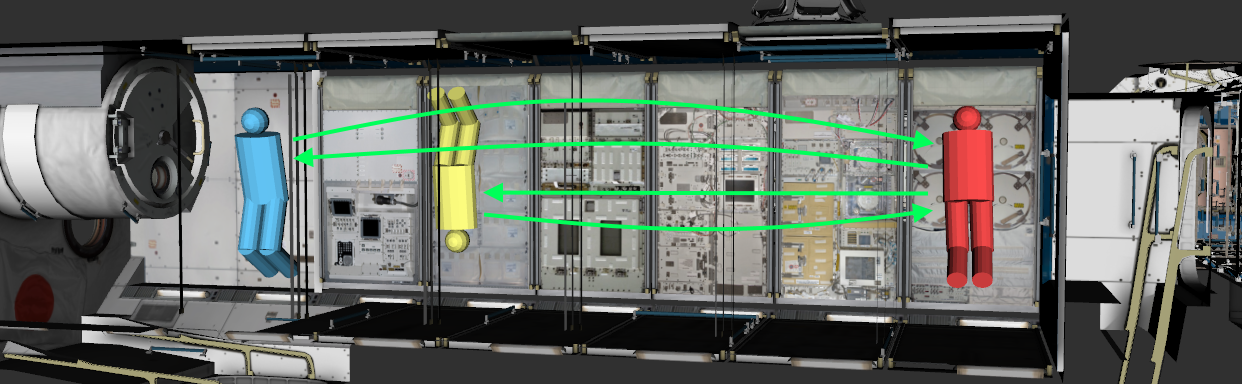}
	\vspace{-.5cm}
	\caption{Equipment transfer task.}
	\vspace{-.1cm}
	\label{fig:transfer}
\end{subfigure}
\caption{Visualization of human task trajectories, with three poses shown to give an example of how the human may move and rotate through the task.}
\label{fig:trajectories}
\vspace{-0.2cm}
\end{figure}

\begin{figure*}[t]
\centering
\includegraphics[width=0.975\textwidth]{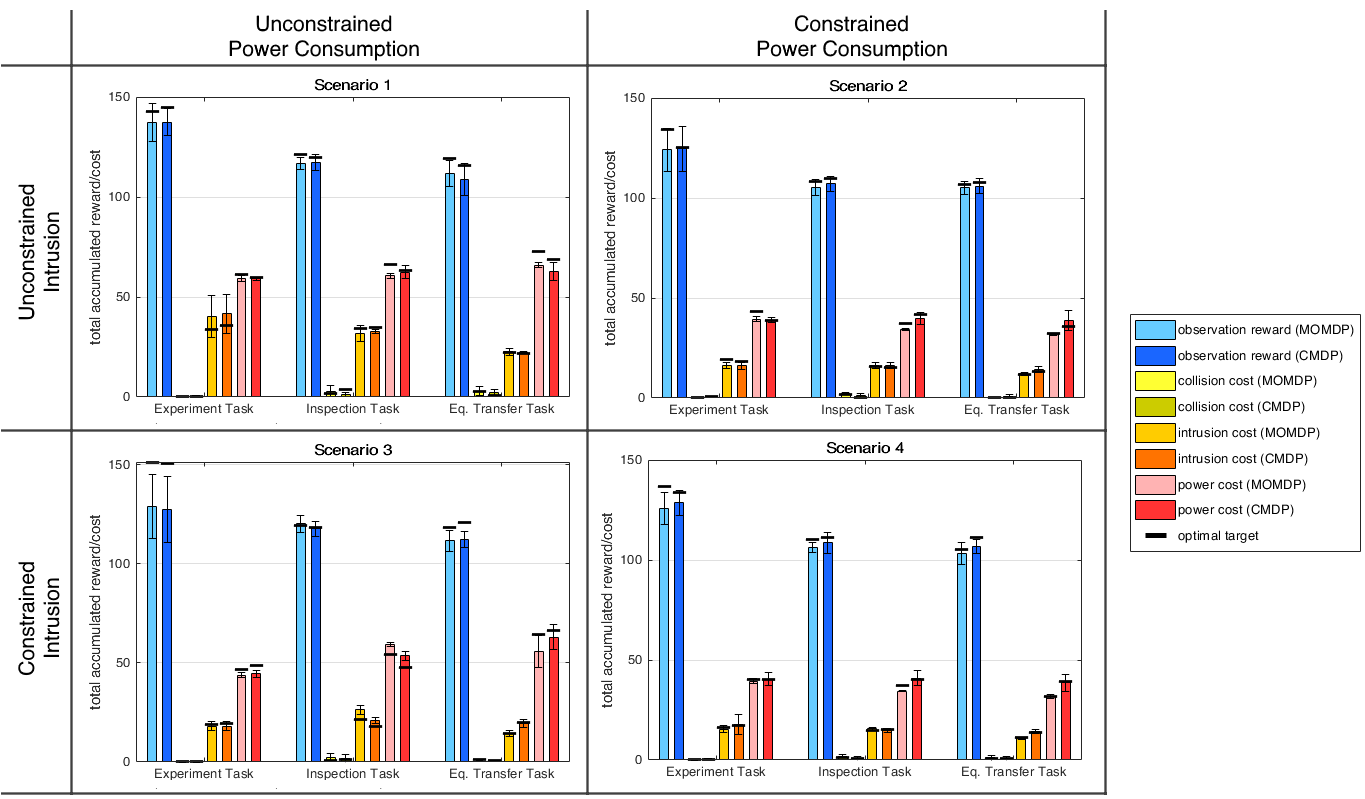}
\caption{Average total accumulated rewards and costs ($\pm 1$ standard deviation) for three human tasks over four cost scenarios.  All rewards and costs are normalized on the interval [0,1] per second, resulting in maximum total rewards and costs of 180 for a 180 second human trajectory.  For context, optimal rewards and costs given an exactly known human trajectory are shown with bold tick marks.}
\label{fig:results}
\vspace{-0.5cm}
\end{figure*}

We evaluate both the MOMDP with linear scalarization and the CMDP method on the Astrobee case study.  We simulate three observation tasks in NASA's Gazebo simulation of the ISS, under four different cost weighting scenarios.  The human's tasks, with trajectories visualized in Figure \ref{fig:trajectories}, are:
\begin{enumerate}
  \item \textit{Experiment}: the human moves between the three experiment stations shown in Figure \ref{fig:experiment}, staying at each pose for an extended duration.
  \item \textit{Inspection}: the human slowly moves over a large area to perform a surface inspection, shown in Figure \ref{fig:inspection}, primarily staying in motion for the duration of the task.
  \item \textit{Equipment transfer}: the human moves back and forth between pick-up points (blue and yellow in Figure \ref{fig:transfer}) and a drop-off point (red in Figure \ref{fig:transfer}), repeatedly moving over long distances.
\end{enumerate}
For each task, we define a trajectory distribution $\mathbf{x}_h(t)$ by fitting Gaussian distributions to both key poses and task timing data, allowing for changes in position, orientation, and task segment speed.  We sample a set of 5 different evaluation trajectories $x_{h\_eval}$ for each task, to ensure that our evaluation is not overfitting to individual trajectories.  The average duration of each task is 180 seconds.

\begin{table}[t]
\caption{Cost weighting scenarios with associated scalarization weights (middle) and constraint thresholds (right).}
\label{tab:weights}
\begin{center}
\renewcommand{\arraystretch}{1.2}
\begin{tabular}{c|l|l}
 & $\mathbf{w}:[ w_{\tilde{r}}, w_{\tilde{c}0}, w_{\tilde{c}1}, w_{c2}]$ & $\mathbf{d}:[d_{c0}, d_{c1}, d_{c2}]$\\
\hline
Scenario 1 & $[0.67, 0.33, 0, 0]$ & $[1, 180, 180]$\\
Scenario 2 & $[0.33, 0.41, 0, 0.26]$ & $[1, 180, 40]$ \\
Scenario 3 & $[0.35, 0.43, 0.22, 0]$ & $[1, 20, 180]$ \\
Scenario 4 & $[0.27, 0.34, 0.17, 0.22]$ & $[1, 20, 40]$ \\
\end{tabular}
\end{center}
\vspace{-0.2cm}
\end{table}

For each evaluation trajectory, we run both the MOMDP and CMDP methods under four different cost weightings, according to the following scenarios:
\begin{itemize}
  \item \textit{Scenario 1}: Avoid collisions, but ignore other costs
  \item \textit{Scenario 2}: Avoid collisions and limit power consumption, being as intrusive as necessary
  \item \textit{Scenario 3}: Avoid collisions and limit intrusiveness, while using as much power as necessary
  \item \textit{Scenario 4}: Reduce collisions, intrusiveness, and power consumption
\end{itemize}
Table \ref{tab:weights} lists scalarization weights and constraint thresholds for each scenario.  Each model is solved over expected reward and costs (see Equations \ref{eq:expr} and \ref{eq:expc}) calculated from $N=10$ trajectories sampled from $\mathbf{x}_h$, with a time step (i.e. a new decision made) of 1 second.  Additionally, as execution is stochastic, the human's trajectory is stochastic, and the optimal CMDP solution is a stochastic policy, we execute 5 runs per solved model per cost weighting scenario.

We evaluate the two methods according to three criteria:
\begin{itemize}
  \item Policy performance with respect to the total accumulated $r$, $c_0$, $c_1$, and $c_2$
  \item Ease of parameter setting for developers, with respect to transparency of their effect on policy performance across tasks
  \item Algorithm runtime
\end{itemize}
Both the MOMDP and CMDP methods produce optimal policies over expected reward and costs for a given set of parameters.  The total accumulated rewards and costs over all evaluation trials are presented in Figure \ref{fig:results}.  For context, we also show the optimal accumulated rewards and costs when solving the model using the exact rewards and costs from $x_{h\_eval}$, as bold tick marks in Figure \ref{fig:results}.  The main takeaway is that when compared to eachother, both algorithms perform comparably to each other.  This is the case for all cost weighting scenarios, including avoiding collisions, constraining intrusiveness, limiting power consumption, and combinations of all three.  Further, in all cases both methods approach the optimal reward and costs for a single known human subject trajectory.

The methods differ significantly in their parameter setting process.  In the case of CMDPs, the effect of parameter setting on performance is much more transparent, as the constraints directly translate to total costs.  The constraints act as budgets on the cost functions (e.g. in Scenario 4, the observation should be performed with at most 20 intrusiveness, using less than 40 power).  In contrast, the MOMDP's scalarization weights represent only relative relationships between rewards and costs, and their effect on total reward and costs depend on the task and environment.  As such the developer must run experimental trials, in simulation or on the real system, to understand exactly what effect a set of weights will produce.  This is especially concerning when the human's task changes, as a set of weights tuned on one set of tasks can produce a different cost profile when run on a new task.

Human decision making literature further highlights the benefits of the CMDP approach over the MOMDP approach with respect to parameter setting.  By requiring acceptable threshold setting, CMDPs reduce the system designer's role to specification of a satisficing problem, whereas by setting weights that combine multiple rewards and costs, MOMDPs with linear scalarization require the designer to perform an optimizing role \cite{eilon1972goals}.  Many studies have shown that, for humans, satisficing takes less time to evaluate than optimizing, and satisficers are less prone to choice paralysis, have lower decision regret, and predict outcomes more accurately \cite{schwartz2002maximizing, parker2007maximizers, jain2013maximizers}.

The algorithms also differ significantly in run time.  The MOMDP with linear scalarization is a simpler model, and as such, backwards induction (Equations \ref{eq:bi1} and \ref{eq:bi2}) solves an MOMDP for a 180 second task in approximately 1 second.  For the same 180 second task, the linear program required to solve a CMDP contains hundreds of thousands of variables, and for our evaluation had a median solve time of 237 seconds\footnote{We use the \texttt{lp\_solve} library to solve the CMDPs' linear programs, with runtimes based on a 3.4 GHz processor.}.  Both methods can be solved a priori when the observation subject's trajectory can be predicted in advance based on previously collected data, and in such cases algorithm run time is not a limitation.  In cases where the robot does not have task data or a reasonable trajectory distribution to sample human trajectories from, only the MOMDP model is fast enough to re-solve in real-time.

Given the above factors, we find that compared to the MOMDP method, CMDPs are more flexible to a variety of tasks and conditions.  The ease of CMDP parameter setting allows a developer to specify zero tolerance for collisions, apply a limit to the degree of intrusion, and set a power budget based on the starting state of the robot, all of which will be respected for any task without the need for testing and tuning.  The CMDP linear program runtime is a significant drawback, however.  While we have shown that the CMDP algorithm can be solved a priori when a task trajectory distribution is available, we note that its long solve time makes the approach unsuitable for cases with greater task or environmental uncertainty, as such cases can require reactive re-planning to account for unexpected human behavior.  In such cases, the computationally efficient MOMDP method is more suitable.

\section{CONCLUSION AND FUTURE WORK}
\label{sec:future}
\vspace{-0.1cm}
In this work, we have presented two methods for solving the autonomous observation problem, and demonstrated their performance for a specific system and 6-DOF environment.  While both methods produce optimal policies, we find that, because of its satisficing methodology, the CMDP formulation is preferable due to its clear relationship between cost threshold setting and policy performance.  Due to their requirement to be solved offline, however, we find that CMDPs are not practical for all real-world tasks.  While we have shown that the CMDP algorithm can be used over task trajectory distributions, which account for \textit{expected} deviations in the human subject's path, an ideal algorithm would be able to account for \textit{unexpected} trajectory deviations by incorporating online re-planning.  The linear scalarization MOMDP approach can be fast enough to solve for updated policies in real-time\footnote{We consider 1 second a reasonable decision period for the Astrobee task.}, but the CMDP's 4 minute run time is orders of magnitude too slow.

While some work has been done in finding any-time approximate solutions to CMDPs using Monte Carlo Tree Search (see the Cost-Constrained UCT algorithm\cite{LKPK2018a}), both the branching factor and search depth of our problem domain are too large for Monte Carlo search to approximate optimal solutions in real-time.  In future work, we plan to explore approaches for approximating CMDP solutions so that the CMDP method can adapt to unexpected trajectory changes in real-world autonomous observation tasks.

\section*{Acknowledgement}
This work is supported in part by an Early Career Faculty grant from NASA’s Space Technology Research Grants Program.

\addtolength{\textheight}{-7.5cm}

\bibliographystyle{IEEEtran}
\bibliography{IEEEabrv,icra2020-kent-chernova}

\begin{thebibliography}{10}
\providecommand{\url}[1]{#1}
\csname url@rmstyle\endcsname
\providecommand{\newblock}{\relax}
\providecommand{\bibinfo}[2]{#2}
\providecommand\BIBentrySTDinterwordspacing{\spaceskip=0pt\relax}
\providecommand\BIBentryALTinterwordstretchfactor{4}
\providecommand\BIBentryALTinterwordspacing{\spaceskip=\fontdimen2\font plus
\BIBentryALTinterwordstretchfactor\fontdimen3\font minus
  \fontdimen4\font\relax}
\providecommand\BIBforeignlanguage[2]{{%
\expandafter\ifx\csname l@#1\endcsname\relax
\typeout{** WARNING: IEEEtran.bst: No hyphenation pattern has been}%
\typeout{** loaded for the language `#1'. Using the pattern for}%
\typeout{** the default language instead.}%
\else
\language=\csname l@#1\endcsname
\fi
#2}}

\bibitem{smith2016astrobee}
T.~Smith, J.~Barlow, M.~Bualat, T.~Fong, C.~Provencher, H.~Sanchez, and
  E.~Smith, ``Astrobee: A new platform for free-flying robotics on the
  international space station,'' 2016.

\bibitem{CIMON}
``{CIMON} - the intelligent astronaut assistant,''
  \url{https://www.dlr.de/dlr/en/desktopdefault.aspx/tabid-10212/332_read-26307/#/gallery/29911},
  published: 2018-03-02.

\bibitem{ishida2012marker}
M.~Ishida and K.~Shimonomura, ``Marker based camera pose estimation for
  underwater robots,'' in \emph{2012 IEEE/SICE International Symposium on
  System Integration (SII)}.\hskip 1em plus 0.5em minus 0.4em\relax IEEE, 2012,
  pp. 629--634.

\bibitem{bajcsy2018revisiting}
R.~Bajcsy, Y.~Aloimonos, and J.~K. Tsotsos, ``Revisiting active perception,''
  \emph{Autonomous Robots}, vol.~42, no.~2, pp. 177--196, 2018.

\bibitem{chen2011active}
S.~Chen, Y.~Li, and N.~M. Kwok, ``Active vision in robotic systems: A survey of
  recent developments,'' \emph{International Journal of Robotics Research},
  vol.~30, no.~11, pp. 1343--1377, 2011.

\bibitem{nilsson2009towards}
U.~Nilsson, P.~{\"O}gren, and J.~Thunberg, ``Towards optimal positioning of
  surveillance ugvs,'' in \emph{Optimization and Cooperative Control
  Strategies}.\hskip 1em plus 0.5em minus 0.4em\relax Springer, 2009, pp.
  221--233.

\bibitem{bodor2005mobile}
R.~Bodor, A.~Drenner, M.~Janssen, P.~Schrater, and N.~Papanikolopoulos,
  ``Mobile camera positioning to optimize the observability of human activity
  recognition tasks,'' in \emph{IEEE/RSJ International Conference on
  Intelligent Robots and Systems}, 2005, pp. 1564--1569.

\bibitem{schroeter2009autonomous}
C.~Schroeter, M.~Hoechemer, S.~Mueller, and H.-M. Gross, ``Autonomous robot
  cameraman-observation pose optimization for a mobile service robot in indoor
  living space,'' in \emph{2009 IEEE International Conference on Robotics and
  Automation}, 2009, pp. 424--429.

\bibitem{ma2009matching}
C.~Y. Ma, D.~K. Yau, J.-c. Chin, N.~S. Rao, and M.~Shankar, ``Matching and
  fairness in threat-based mobile sensor coverage,'' \emph{IEEE Transactions on
  Mobile Computing}, vol.~8, no.~12, pp. 1649--1662, 2009.

\bibitem{best2016multi}
G.~Best, J.~Faigl, and R.~Fitch, ``Multi-robot path planning for budgeted
  active perception with self-organising maps,'' in \emph{IEEE/RSJ
  International Conference on Intelligent Robots and Systems (IROS)}, 2016, pp.
  3164--3171.

\bibitem{arora2017randomized}
S.~Arora and S.~Scherer, ``Randomized algorithm for informative path planning
  with budget constraints,'' in \emph{IEEE International Conference on Robotics
  and Automation (ICRA)}, 2017, pp. 4997--5004.

\bibitem{wang2008awareness}
Y.~Wang, I.~Hussein, and R.~S. Erwin, ``Awareness-based decision making for
  search and tracking,'' in \emph{American Control Conference}.\hskip 1em plus
  0.5em minus 0.4em\relax IEEE, 2008, pp. 3169--3175.

\bibitem{sung2017algorithm}
Y.~Sung and P.~Tokekar, ``Algorithm for searching and tracking an unknown and
  varying number of mobile targets using a limited fov sensor,'' in \emph{IEEE
  International Conference on Robotics and Automation (ICRA)}, 2017, pp.
  6246--6252.

\bibitem{popovic2017online}
M.~Popovi{\'c}, G.~Hitz, J.~Nieto, I.~Sa, R.~Siegwart, and E.~Galceran,
  ``Online informative path planning for active classification using uavs,'' in
  \emph{IEEE international conference on robotics and automation (ICRA)}, 2017,
  pp. 5753--5758.

\bibitem{garrell2017aerial}
A.~Garrell, L.~Garza-Elizondo, M.~Villamizar, F.~Herrero, and A.~Sanfeliu,
  ``Aerial social force model: A new framework to accompany people using
  autonomous flying robots,'' in \emph{IEEE/RSJ International Conference on
  Intelligent Robots and Systems (IROS)}, 2017, pp. 7011--7017.

\bibitem{ragi2013uav}
S.~Ragi and E.~K. Chong, ``Uav path planning in a dynamic environment via
  partially observable markov decision process,'' \emph{IEEE Transactions on
  Aerospace and Electronic Systems}, vol.~49, no.~4, pp. 2397--2412, 2013.

\bibitem{chen2016pomdp}
M.~Chen, E.~Frazzoli, D.~Hsu, and W.~S. Lee, ``Pomdp-lite for robust robot
  planning under uncertainty,'' in \emph{IEEE International Conference on
  Robotics and Automation (ICRA)}, 2016, pp. 5427--5433.

\bibitem{roijers2013survey}
D.~M. Roijers, P.~Vamplew, S.~Whiteson, and R.~Dazeley, ``A survey of
  multi-objective sequential decision-making,'' \emph{Journal of Artificial
  Intelligence Research}, vol.~48, pp. 67--113, 2013.

\bibitem{eck2012evaluating}
A.~Eck and L.-K. Soh, ``Evaluating pomdp rewards for active perception,'' in
  \emph{International Conference on Autonomous Agents and Multiagent
  Systems-Volume 3}.\hskip 1em plus 0.5em minus 0.4em\relax International
  Foundation for Autonomous Agents and Multiagent Systems, 2012, pp.
  1221--1222.

\bibitem{lizotte2012linear}
D.~J. Lizotte, M.~Bowling, and S.~A. Murphy, ``Linear fitted-q iteration with
  multiple reward functions,'' \emph{Journal of Machine Learning Research},
  vol.~13, no. Nov, pp. 3253--3295, 2012.

\bibitem{freire2017gubs}
V.~Freire and K.~V. Delgado, ``Gubs: A utility-based semantic for goal-directed
  markov decision processes,'' in \emph{Conference on Autonomous Agents and
  MultiAgent Systems}.\hskip 1em plus 0.5em minus 0.4em\relax International
  Foundation for Autonomous Agents and Multiagent Systems, 2017, pp. 741--749.

\bibitem{ding2014hierarchical}
X.~D. Ding, B.~Englot, A.~Pinto, A.~Speranzon, and A.~Surana, ``Hierarchical
  multi-objective planning: From mission specifications to contingency
  management,'' in \emph{IEEE international conference on robotics and
  automation (ICRA)}, 2014, pp. 3735--3742.

\bibitem{feyzabadi2016multi}
S.~Feyzabadi and S.~Carpin, ``Multi-objective planning with multiple high level
  task specifications,'' in \emph{IEEE International Conference on Robotics and
  Automation (ICRA)}, 2016, pp. 5483--5490.

\bibitem{hu2007markov}
Q.~Hu and W.~Yue, \emph{Markov decision processes with their
  applications}.\hskip 1em plus 0.5em minus 0.4em\relax Springer Science \&
  Business Media, 2007, vol.~14.

\bibitem{edward1966hall}
E.~T. Hall, ``The hidden dimension,'' 1966.

\bibitem{altman1999constrained}
E.~Altman, \emph{Constrained Markov decision processes}.\hskip 1em plus 0.5em
  minus 0.4em\relax CRC Press, 1999, vol.~7.

\bibitem{eilon1972goals}
S.~Eilon, ``Goals and constraints in decision-making,'' \emph{Journal of the
  Operational Research Society}, vol.~23, no.~1, pp. 3--15, 1972.

\bibitem{schwartz2002maximizing}
B.~Schwartz, A.~Ward, J.~Monterosso, S.~Lyubomirsky, K.~White, and D.~R.
  Lehman, ``Maximizing versus satisficing: Happiness is a matter of choice.''
  \emph{Journal of personality and social psychology}, vol.~83, no.~5, p. 1178,
  2002.

\bibitem{parker2007maximizers}
A.~M. Parker, W.~B. De~Bruin, and B.~Fischhoff, ``Maximizers versus
  satisficers: Decision-making styles, competence, and outcomes,''
  \emph{Judgment and Decision making}, vol.~2, no.~6, p. 342, 2007.

\bibitem{jain2013maximizers}
K.~Jain, J.~N. Bearden, and A.~Filipowicz, ``Do maximizers predict better than
  satisficers?'' \emph{Journal of Behavioral Decision Making}, vol.~26, no.~1,
  pp. 41--50, 2013.

\bibitem{LKPK2018a}
J.~Lee, G.-H. Kim, P.~Poupart, and K.-E. Kim, ``Monte-carlo tree search for
  constrained mdps,'' in \emph{ICML/IJCAI/AAMAS Workshop on Planning and
  Learning (PAL)}, 7 2018.

\end{thebibliography}






\end{document}